# Automatic Text Line Segmentation Directly in JPEG Compressed Document Images

**Bulla Rajesh, Mohammed Javed, P. Nagabhushan**

Department of IT, Indian Institute of Information Technology, Allahabad, U.P -211012, India
{Email: *javed@iiita.ac.in}

**Abstract:** JPEG is one of the popular image compression algorithms that provide efficient storage and transmission capabilities in consumer electronics, and hence it is the most preferred image format over the internet world. In the present digital and Big-data era, a huge volume of JPEG compressed document images are being archived and communicated through consumer electronics on daily basis. Though it is advantageous to have data in the compressed form on one side, however, on the other side processing with off-the-shelf methods becomes computationally expensive because it requires decompression and recompression operations. Therefore, it would be novel and efficient, if the compressed data are processed directly in their respective compressed domains of consumer electronics. In the present research paper, we propose to demonstrate this idea taking the case study of printed text line segmentation. Since, JPEG achieves compression by dividing the image into non overlapping 8×8 blocks in the pixel domain and using Discrete Cosine Transform (DCT); it is very likely that the partitioned 8×8 DCT blocks overlap the contents of two adjacent text-lines without leaving any clue for the line separator, thus making text-line segmentation a challenging problem. Two approaches of segmentation have been proposed here using the DC projection profile and AC coefficients of each 8×8 DCT block. The first approach is based on the strategy of partial decompression of selected DCT blocks, and the second approach is with intelligent analysis of F10 and F11 AC coefficients and without using any type of decompression. The proposed methods have been tested with variable font sizes, font style and spacing between lines, and a good performance is reported.

## I. Introduction

In today's digital era of consumer electronics, a huge volume of document images are being archived and communicated in the compressed form for the efficiency of storage in databases and transmission over the internet (social media, Email, E-Governance applications, etc.). Based on the requirement of handling large volume of multimedia data, various compression algorithms have been proposed in the literature [1], [2]. However, JPEG is the most preferred compressed file format in the consumer electronics and internet world, and more than 90% of images in internet databases are JPEG compressed images [3]. So, developing algorithms to process compressed data would be novel and efficient as already reported in the literature [2]. The segmentation work in JPEG by [4] is not fully implemented in compressed domain. So, we propose to carry out printed text-line segmentation directly in JPEG compressed document images. It is significant to develop text line segmentation for JPEG documents, since many consumer electronic applications like OCR, word spotting, and layout extraction can be attempted based on this work.

## II. Problem Description

JPEG algorithm divides image into 8×8 non-overlapping blocks in pixel domain, and subsequently applies DCT for each block. During this process, it is very likely that 8×8 blocks may overlap the characters, words and the entire text from both rows of any two adjacent text lines, and further it may not provide any indication for the existence of a line separator between them. The absence of clear line separator between the adjacent text lines makes printed text line segmentation in JPEG compressed document images, a challenging task (see Fig.1). In any document the font size of English alphabet varies, with different styles and space. In such a scenario, a 8×8 DCT partitioning block may contain text information in various forms as shown in Fig. 2. In the Fig. 1, we observe that, between 1st and 2nd printed text lines there is a very less space, between 2nd and 3rd text lines there is sufficient space of one 8×8 block, and between 3rd and 4th text lines there is no space at all. Here each block is of size 8×8 and during DCT compression it cannot be exactly predicted how the text and spacing between the lines will be encoded. Therefore, the problem has to be intelligently analysed and a strategy for text line segmentation has to be devised to take care of all the cases shown in Fig. 1.

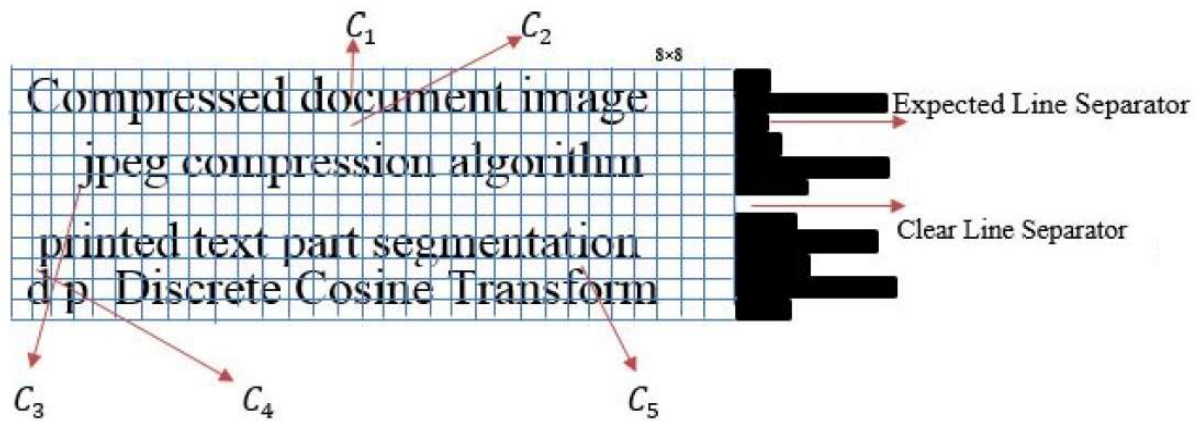

Fig. 1. Illustration of how a 8×8 DCT block encodes printed text contents in pixel domain in different scenarios.

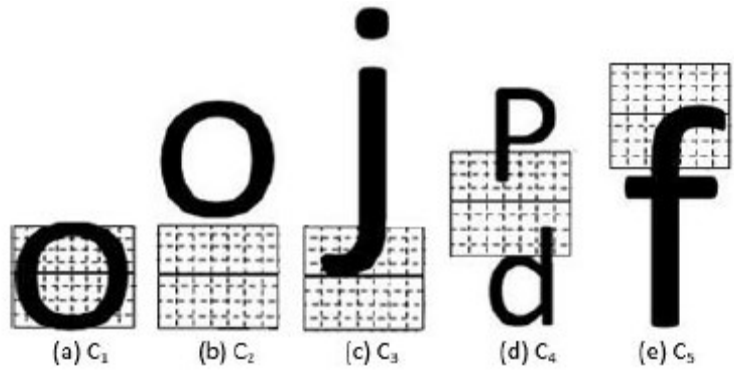

Fig. 2. Depicting the various possible blocks shown in Fig .1.

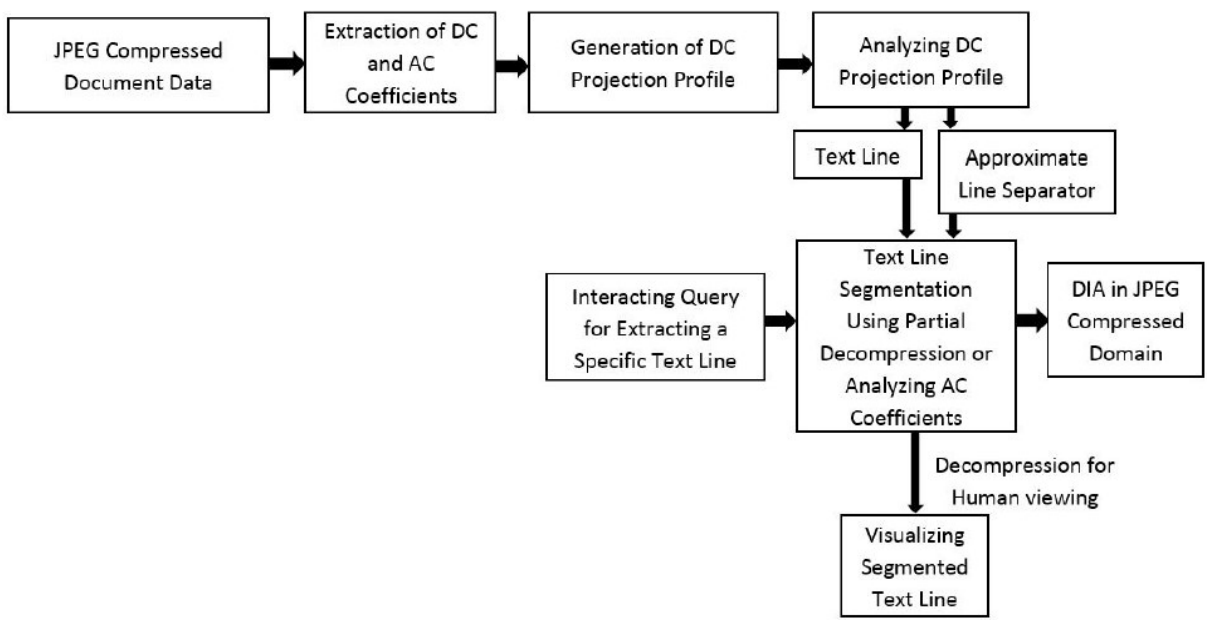

Fig. 3. Proposed model for segmentation in JPEG compressed documents.

In practical scenarios, a document may or may not have enough spacing between the text lines. Based on this, we categorize a 8×8 DCT block into five categories as shown in Fig. 2. Fig. 2 is reproduced form Fig. 1. Except in case of (b), every 8×8 DCT block needs to be analysed for the presence of text content and the decision about line separator has to be made. During DCT compression, partitioning of 8×8 blocks for bigger letters and small letters are different. For example in Fig. 2, partitioning in case of letter 'j' is observed in the below portion. Whereas, for letter 'f' partitioning is seen in the top portion. For letter 'o', block partitioning is perfectly within the standard block size. Therefore, just by looking at the DCT blocks of base region of the text line, decision about the text line boundary cannot be made. So, every block above and below the base region of text-line needs to be analysed and decision about its relationship with current text-line and adjacent text-line has to be taken.

# III. Proposed Model

Different stages involved in the proposed model are shown in Fig. 3. Two strategies are devised. The first strategy is based on DC projection profile and partial decompression of selected DCT blocks. The second strategy is based on DC projection profile and intelligent analysis of AC coefficients F10 and F11. The block diagram for both the models follow same sequence of steps as shown in Fig. 3. In a typical 8×8 DCT block, the DC coefficient represents the average information within the block, and if the 8×8 block passes through text, then it will be reflected in the DC coefficient. From the JPEG compressed stream, we therefore extract all the DC coefficients and add them horizontally to get vertical DC projection profile. In the profile, if $DC_i > 0$ implies the presence of text and $DC_i = 0$ hints at presence of line separator. However, in many cases, the line separator may not be clearly visible due to spacing and font constraints. In such case the line separator needs to be predicted using the peak of DC projection profile that will be obtained at base region of text line see Fig. 1. Mid-block between any two peaks is taken as expected line separator and then two proposed strategies are applied to detect the exact boundary of line.

### *A. Partial Decomposition of selected blocks*

Once, the expected text line separators are detected using the DC projection profile, next it decompress the blocks along the expected text-line separator to find out boundary of the text-line in the pixel domain through vertical projection profile analysis.

### *B. Using AC Coefficients of selected blocks*

If we closely observe the AC coefficients of any 8×8 DCT block, they are linear combination of all the 64 pixels in that block. If the 8×8 pixel block contains the cases (c) and (e) shown in Fig. 2, then by using F10 coefficient we can resolve the text line boundary. In printed text lines, the text characters would be horizontally and vertically aligned. It can be seen that, the big sized letters occupy either first portion of rows or last last portion of rows of a 8×8 block. Therefore to resolve such a DCT block and to separate text lines we simply check F10 and F11 AC coefficients from 8×8 DCT block. If F10> 0 then that character belongs to the above text line and such block will become part of exact boundary of the text line and F11 is used for further analysis.

TABLE I : TEXT-LINE EXTRACTION RESULTS OF BOTH THE METHODS

| Image | P1(%) | R1(%) | F1(%) | P2(%) | R2(%) | F2(%) |
|---|---|---|---|---|---|---|
| 96 | 100 | 100 | 100 | 100 | 100 | 100 |
| 200 | 100 | 100 | 100 | 100 | 90 | 94.73 |
| 300 | 90 | 90 | 90 | 100 | 90 | 94.73 |

TABLE II : EXECUTION TIME ANALYSIS OF THE PROPOSED JPEG COMPRESSED DOMAIN ALGORITHMS WITH RESPECT TO PIXEL DOMAIN PROCESSING.

| Approaches | Segmentation time(sec) | improvement in speed(%) |
|---|---|---|
| Pixel Domain (Decompression + Processing) | 0.53135 | - |
| Partial Decompression | 0.00995 | 98.12 |
| Using AC Coefficients | 0.0006078 | 99.88 |

## IV. Experimental Results

We collected around 100 JPEG compressed document images from various sources of print media like single column journal and conference papers, text books and magazines. Overall, we created 300 images with 100 documents under each category of 96 dpi, 200 dpi, and 300 dpi. The experimental results using both the proposed approaches are reported in in Table I (where, P-Precision, R-Recall, F-F-measure). We found that our proposed algorithms give good performance and the speed gained are reported in Table II.

## V. Conclusion

We showed two strategies to segment text lines directly from JPEG compressed documents. First is using decompression of selected blocks along with expected line separator and second is directly in the compressed domain, using AC (F10 & F11) coefficients. Although dealing directly with compressed data is challenging but we observed that our strategies are working faster than other traditional approaches.

## References

[1] J. Mukhopadhyay, Image and video processing in the compressed domain. Chapman and Hall/CRC, 2011.
[2] M. Javed, P. Nagabhushan, and B. B. Chaudhuri, "A review on document image analysis techniques directly in the compressed domain," Artificial Intelligence Review, vol. 50, no. 4, pp. 539–568, 2018.
[3] C. Florea, M. Gordan, B. Orza, and A. Vlaicu, "Compressed domain computationally efficient processing scheme for jpeg image filtering," in Advanced Engineering Forum, vol. 8, pp. 480–489, Trans Tech Publ, 2013.
[4] R. L. de Queiroz and R. Eschbach, "Fast segmentation of the JPEG compressed documents," Journal of Electronic Imaging, vol. 7, no. 2, pp. 367–378, 1998.